\documentclass{article}

\PassOptionsToPackage{numbers,compress,sort}{natbib}

\usepackage{enumitem}

\usepackage[final]{neurips_2024}

\usepackage[utf8]{inputenc} 
\usepackage[T1]{fontenc}    
\usepackage{hyperref}       
\usepackage{url}            
\usepackage{booktabs}       
\usepackage{amsfonts}       
\usepackage{nicefrac}       
\usepackage{microtype}      
\usepackage{xcolor}         

\usepackage{soul}
\usepackage{amsmath}
\usepackage{mathtools}
\usepackage{amssymb}
\usepackage{caption}
\usepackage{subcaption}
\usepackage{wrapfig}
\usepackage{dsfont}

\newcommand{\bx}{\boldsymbol{x}}
\newcommand{\bv}[1]{\mathbf{#1}} 

\newcommand{\ind}{\mathds{1}}

\DeclareMathOperator*{\argmax}{arg\,max}

\title{Dynamic layer selection in decoder-only transformers}

\author{%
  Theodore Glavas$^{123}$\footnotemark[1] \quad Joud Chataoui$^{123}$\footnotemark[1] \quad Florence Regol$^{123}$ \quad Wassim Jabbour$^{1}$ \\
  \textbf{Antonios Valkanas}$^{123}$ \quad \textbf{Boris N. Oreshkin}$^{4}$\footnotemark[2] \quad \textbf{Mark Coates}$^{123}$ \\
  $^1$McGill University \quad $^2$International Laboratory on Learning Systems (ILLS) \\
  $^3$Mila Québec AI Institute \quad $^4$Amazon Science\\
  \texttt{\{theodore.glavas, joud.chataoui} \\
  \texttt{florence.robert-regol, wassim.jabbour, antonios.valkanas\} @mail.mcgill.ca} \\
  \texttt{boris.oreshkin@gmail.com, mark.coates@mcgill.ca}
}

\begin{document}

\maketitle
\footnotetext[1]{Equal contribution}
\footnotetext[2]{This work is not related to the author’s position at Amazon.}

\begin{abstract}
The vast size of Large Language Models (LLMs) has prompted a search to optimize inference. One effective approach is dynamic inference, which adapts the architecture to the sample-at-hand to reduce the overall computational cost. We empirically examine two common dynamic inference methods for natural language generation (NLG): layer skipping and early exiting. We find that a pre-trained decoder-only model is significantly more robust to layer removal via layer skipping, as opposed to early exit. We demonstrate the difficulty of using hidden state information to adapt computation on a per-token basis for layer skipping. Finally, we show that dynamic computation allocation on a per-sequence basis holds promise for significant efficiency gains by constructing an oracle controller. Remarkably, we find that there exists an allocation which achieves equal performance to the full model using only $23.3\%$ of its layers on average.

\end{abstract}
\vspace{-0.5cm}
\section{Introduction}
\vspace{-0.2cm}

The distribution and reuse of pre-trained LLMs means that inference now dominates the computational cost over the model's lifecycle \citep{sustainable_ai_mlsys, power_hungry}. As such, a variety of techniques have been proposed to improve LLM inference efficiency~\cite{quant_survey, dnn_survey}. While some approaches such as quantization and structural pruning can be readily applied with minor modifications, the auto-regressive nature of natural language generation (NLG) imposes new challenges for input-adaptive techniques such as dynamic networks. \\
\textbf{First}, the generation output is highly sensitive to error introduced by reducing computation for any one token. Interdependencies cause error to compound by (a) relying on incorrect previously predicted tokens, and (b) filling missing KV cache entries with approximations \citep{depth_adaptive_transformer_elbayad,calm_schuster}. Despite some empirical studies \citep{deja_vu_liu,layerdrop_fan}, the existing literature fails to unequivocally answer whether early exit is better than layer skipping at minimizing the error in the model's hidden states.\\
\textbf{Second}, the large output vocabulary space in NLG renders the evaluation of the classification head at intermediate layers prohibitively costly. Instead, lightweight controllers route a token using the hidden state~\cite{depth_adaptive_transformer_elbayad, calm_schuster, zeng2023_learning_to_skip}. It is not clear that existing token-level layer skipping controllers derive a genuine benefit from processing the hidden state. \\
\textbf{Third}, per-token adaptation adds overhead to each autoregressive pass, which accumulates over long generations. Per-token controllers must therefore be lightweight networks to mitigate this effect \citep{depth_adaptive_transformer_elbayad,calm_schuster,zeng2023_learning_to_skip,consistent_ee_zeng}. We investigate in Section \ref{sec:oracle-layer-skipping} whether it is possible to reduce computation, with a limited performance drop, by making sequence-level decisions that only need to be evaluated once.\\
In this work, we conduct a series of three experiments and making the following contributions\footnote[3]{The experiment code is available  \href{https://github.com/networkslab/enlsp_neurips24}{on this Github repository}.}.:
\begin{enumerate}[leftmargin=*]
  \setlength\itemsep{0em}
    \item We compare the early exit and layer skip mechanisms on a fixed model. Our results show that layer skipping is the preferable strategy for minimizing the accumulated hidden state error.
    \item We examine the efficacy of per-token layer skipping  based on the hidden state, using a low-cost controller, and show that learning a token-agnostic skipping probability is equally effective.
    \item By introducing an oracle controller that optimally assigns computation to sequences, we show that per-sequence adaptation has the potential to achieve a dramatic computation reduction.
\end{enumerate}

\section{Problem Setting}
We consider the problem of auto-regressive language generation using a neural network, denoted by $T(\cdot) = d^L \circ \dots \circ d^1 \circ e(\cdot)$, and consisting of stacked decoder blocks $d^l$ and an embedding layer $e$. For NLG, we augment $T$ with a classification head $\sigma$. Consider an input sequence $\bv{s}_{t} = [\bx_{1: p}, \bv{y}_{1 : t}]$ consisting of the concatenation of a prompt $\bx_{1: p}$ and a sequence of generated tokens $\bv{y}_{1 : t}$  where $x_k, y_k \in \mathcal{Y}$, with $\mathcal{Y}$ denoting the vocabulary space. 
Sequences are generated auto-regressively by sampling the next token $\hat{y}_{t + 1}$ from the predicted probability distribution $\bv{\hat{p}}_{t + 1} = \sigma( T(\bv{s}_{t}))$. In the field of efficient dynamic inference, the early exit method dynamically selects an exit layer $L_{E}$, and executes layers $\{d^l\}_{l \leq L_E}$. Thus, the network can be written as $T_{EE}(\cdot) = d^{L_{E}} \circ \dots \circ d^1 \circ e(\cdot)$. For layer skipping, the execution of each layer $d^l$ is controlled by a binary variable $G^l$, where $G^l = 1$ denotes executing layer $l$. The layer skipping model output is then defined as $T_{LS}(\cdot) = \left(G^L d^L + (1 -G^L) I \right) \circ \dots \circ \left(G^1 d^1 + (1 -G^1) I\right) \circ e(\cdot)$, where $I$ is the identity function. For each case, we define the computational cost $c$ as the number of decoder layers executed, omitting the fixed costs of the embedding layer $e(\cdot)$ and classification head $\sigma(\cdot)$. For early exiting, $c = L_E$, while $c = \sum_{l=1}^L G_l$ for layer skipping. For all experiments, we only consider the computation performed on the autoregressively generated tokens $\bv{y}_{1:T}$, as is commonly done \citep{depth_adaptive_transformer_elbayad, calm_schuster, unified_layer_skipping_liu, skipdecode_delcorro}. The prompt is propagated through the full network since the prompt tokens can be efficiently processed once and in parallel and provide important context for NLG \citep{skipdecode_delcorro}. In general, we aim to achieve the best ROUGE metric \citep{lin-2004-rouge}  for the lowest computational cost $c$.

\section{Empirical Studies, Results, and Analysis}

\subsection{Is layer skipping more effective than early exiting?} \label{sec:Comparing-EE-LS}

We hypothesize that for NLG tasks and a pre-trained model without fine-tuning, layer skipping is more effective than early exit. Previous empirical results show that residual connections in decoder blocks contribute a greater portion of the total block's output than the MLP and attention blocks~\citep{deja_vu_liu}. Consequently, skipping a layer's execution should result in limited hidden state drift. We assess ``effectiveness'' by two methods: (i) for next-token prediction error, we assess how the final-layer hidden state is preserved for a specific computational cost, where preservation is measured via cosine similarity denoted $\mathcal{S}(\bv{h}_t^L, \bv{h}^L_{t, c})$; (ii) for KV cache propagation error, we measure the layerwise cosine similarity of all layers' hidden states used for KV caching, and report the mean.  

\paragraph{Experiment setup:} Using OPT-1.3B \citep{zhang2022opt} fine-tuned on Alpaca \citep{alpaca}, and given a computational cost $c \in \{1, \dots, L\}$, we compare the early exiting $T_\text{EE,c}$ (where $L_E = c$) with two layer skipping approaches. The {\em uniform} layer skipping network $T_{\text{ULS, c}}$ evenly spaces skipped and executed layers~\citep{unified_layer_skipping_liu}. The {\em random} layer skipping network $T_{\text{RLS, c}}$ executes $c$ random layers, shuffling the executed layers every batch. Layer $l = 1$ always executes, since it plays a important role in transforming embeddings~\citep{deja_vu_liu}. Furthers details and analysis are provided in Appendices~\ref{appendix:datasets_and_models} and~\ref{appendix:hidden_state_similarity}.  

\begin{figure}[h!]
   \centering
   
    \begin{subfigure}{0.5\textwidth}
     \centering
    \includegraphics[scale=0.45]{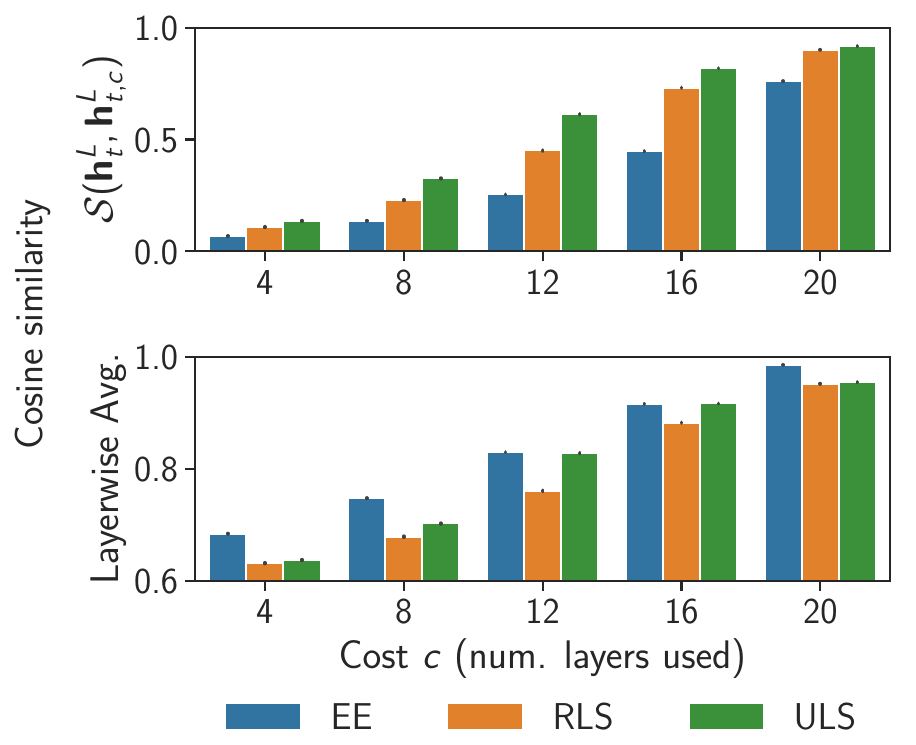}
    \caption{Hidden state cosine similarity.}
    \label{fig:cosine_similarity_vs_computational_budget}
    \end{subfigure}%
    \begin{subfigure}{0.5\textwidth}
    \centering
    \includegraphics[scale=0.45]{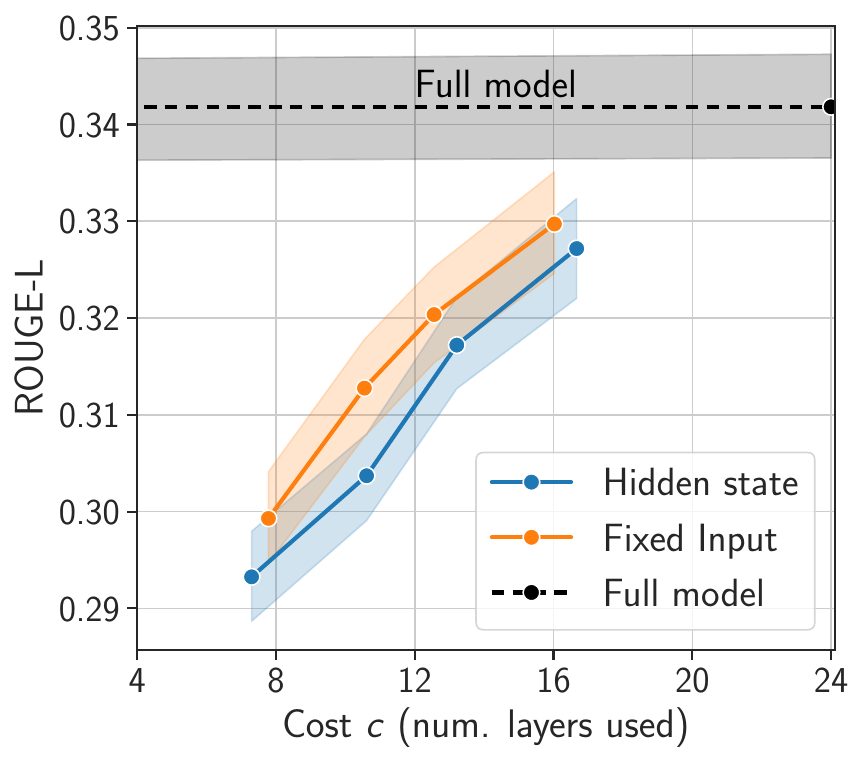}
    \caption{Assessing the performance of hidden state.}
    \label{fig:hs_gumbels}
    \end{subfigure}
    \caption{Using OPT-1.3B (24 layers) on the Alpaca dataset. We report the mean and 95\% confidence interval on the test set. \textbf{(a):} Average cosine similarity of the final hidden states $\bv{h}^L_{t}$ (top) and layerwise intermediate hidden states $\bv{h}^l_{t}, 1\leq l \leq 23$ (bottom) obtained from early exit (EE), random and uniform layer skipping (RLS/ULS) compared to the full model execution. \textbf{(b)} Comparison of cost-performance curves when training skip controllers to use the hidden states $\bv{h}^{l-1}$ or a fixed input.We see that, with statistical significance, using the hidden state is not helpful to the model's performance. }
    \label{fig:exp1_2}
\end{figure}

\paragraph{Results:} Figure \ref{fig:cosine_similarity_vs_computational_budget} shows that the final layer hidden state obtained from uniform layer skipping $\bv{h}^L_{t, \text{ULS, c}}$ is closest to the original hidden state $\bv{h}^L_{t}$. Early exit has the lowest final hidden state similarity, but the highest overall cosine similarity of intermediate hidden states. These results reveal an interesting trade-off: Early exit concentrates error into the upper layers, leading to poorer next-token prediction, but improving average KV cache accuracy. Conversely, skipping layers more evenly distributes hidden state error, improving token prediction despite less accurate caches. At costs $c=\{12,16\}$, uniform layer skipping is a clear winner, matching early exit for similarity of intermediate hidden states while achieving a much more similar final layer state.

\subsection{Are layer skip controllers using hidden states more effective than a constant input?} \label{sec:hidden_state_evaluation}
We hypothesize that layer skipping controllers, limited in capacity because of their associated overhead costs, are ineffective at extracting meaningful context from token hidden states, since they are high dimensional embeddings designed to be processed by much larger decoder blocks. We assess ``effectiveness'' by comparing the performance to a sample-agnostic baseline, wherein the same controllers are provided with a constant input instead of the hidden state.

\paragraph{Experiment formulation:}
Following the approach in \cite{zeng2023_learning_to_skip}, we start by augmenting a fine-tuned model $T$ with $L$ trainable skip controllers to model the skip variable $G^l := g_{\phi^l}^l(\bv{h}^{l-1}_{t})$ at each layer with  $\bv{h}^{0}_{t} = e(\bv{s}_t)$. The skip-controllers are trained using the following loss formulation:
\begin{align}\label{eq:ce_loss}
    \mathcal{L}   &= \mathbb{E}_{\hat{Y}_t, Y_t} [\mathbb{E}_{\bv{G} | Y}[\ind(\hat{Y}_t \neq  Y_t) + \alpha C(\hat{Y}_{t})]] = \mathbb{E}_{\hat{Y}_t, Y} [\mathbb{E}_{\bv{G} | Y}[\ind(\hat{Y}_t \neq  Y_t) + \alpha \sum_{l = 1}^{L}  P(G^l | \bv{s}_{t - 1}; \phi)]]\nonumber \\
    &\approx \frac{1}{N} \sum_{i = 1}^N \Big( KL(\hat{\bv{p}}_{\text{i, t}} \mid \bv{p}_{\text{i, t}}) + \alpha \sum_{l = 1}^{L}  g^l_{\phi}(\bv{h}^{l-1}_{\text{i, t}}) \Big)
\end{align}
The auxiliary term $C(\hat{Y}_{t})$, weighted by $\alpha$, introduces an inference cost penalty proportional to the number of executed layers, providing inference cost regularization. To make the loss differentiable, we use the reparametrization trick introduced in \citet{gumbel_softmax_repar}, which uses the discrete realization $g^l \in \{0, 1\}$ of $G^l$ in the forward path, but uses the continuous $P(G^l | \bv{s}_t; \phi)$ during backpropagation.

\paragraph{Experiment setup:} We model the skip controllers as a single linear layer following \cite{zeng2023_learning_to_skip,calm_schuster,depth_adaptive_transformer_elbayad}, with input size $|\bv{h}|$ and output size $2$. A Gumbel-Softmax module \citep{gumbel_softmax_repar} from the Torch library \citep{pytorch} is attached at the output to allow for differentiable layer skipping \citep{zeng2023_learning_to_skip}. We train the skip controllers jointly with the backbone (we repeat the same experiment with a frozen backbone in Appendix \ref{appendix:ablation_hidden_state}). As a comparison, we train the same network by providing a fixed input tensor of size $|\bv{h}|$ and value 1 as the controller input,  $g_{\phi^l}^l(\bv{1})$. The controller's decision is then no longer sample-aware, but can still learn to optimize the skipping frequency of its associated layer. The output remains stochastic because the Gumbel-Softmax module samples the skip decision from a combination of the linear layer output and a Gumbel(0,1) distribution \citep{gumbel_softmax_repar}. We repeat the procedure varying the hyperparameter $\alpha \in \{2,4,6,10\}$ to obtain different operating points in terms of computational cost. All models are trained using OPT-1.3B on Alpaca. The detailed setup can be found in Appendices~\ref{appendix:datasets_and_models} and~\ref{appendix:training_setup_hidden_states}.

\paragraph{Results:} Figure \ref{fig:hs_gumbels} shows the cost-performance trade-off obtained with and without the hidden states. Since the hidden state does not bring a statistically significant advantage, we conclude that, if we are constrained to use simple skip controllers with limited capacity due to computational restrictions, it is equally effective to learn a token-agnostic average skip rate for each layer. Appendix \ref{appendix:extra_hidden_versus_fixed} compares the average skip ratios of the hidden state and fixed input approaches on a per-layer basis for different $\alpha$ values, showing they follow a similar distribution.

\subsection{Is it possible for sequence-aware layer skipping to outperform a static method?} \label{sec:oracle-layer-skipping}

Considering the outperformance of layer skipping (Sec.~\ref{sec:Comparing-EE-LS}), and the difficulty of using the hidden state for per-token decisions (Sec.~\ref{sec:hidden_state_evaluation}), we assess whether adaptive layer skipping can be performed at the sequence-level. This is further motivated by the following observations: 1) The primary source of latency when auto-regressively generating tokens using an LLM is the repeated transfer of layer weights in and out of GPU memory \cite{spec_decoding_survey, leviathan2023_spec_decoding}. Using a constant subset of layers for an entire sequence generation can thus significantly speed-up inference by reducing the number of IO operations. 2) The controllers need only be evaluated once for an entire generation as opposed to repeatedly for every token. We hypothesize that some prompts are easier to respond to than others and thus need less computation. Therefore, determining the computational path for the whole sequence based on the prompt could provide an efficiency improvement over a static scheme.

\paragraph{Experiment setup:}
\begin{wrapfigure}{r}{0.5\textwidth}\vspace{-2em}
\includegraphics[width=0.95\linewidth]{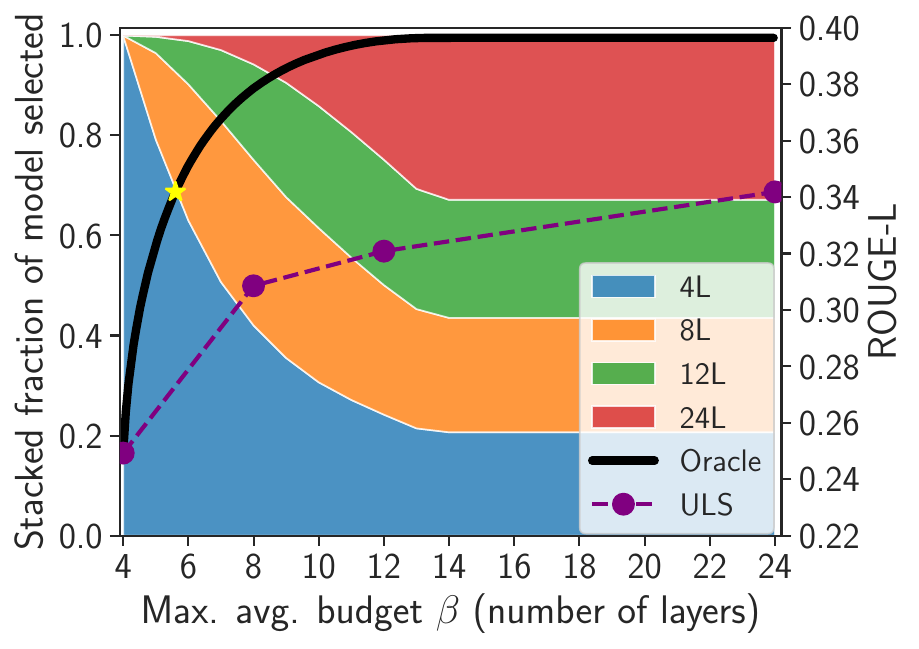}
\caption{ROUGE-L (right axis) of  $T$ and $T_\text{ULS, x}$ compared to an oracle skip controller that selects the optimal model per sequence given a global budget. For each budget, the selection percentage of each model by the oracle is shown as a stacked plot (left axis). The yellow star denotes where the oracle can match the performance of the full model, using an average of 5.6/24 layers ($23.3\%$).} \label{fig:oracle_stacked}
\vspace{-2em}
\end{wrapfigure}
We conduct an oracle experiment to demonstrate the potential of sequence-level methods. The oracle selects the optimal interleaving computational paths for generating multiple tokens. The oracle is given the full OPT-1.3B 24 layer model $T$, as well as 3 models $T_\text{ULS, 12}$, $T_{\text{ULS, 8}}$, and $T_{\text{ULS, 4}}$ that implement uniform layer skipping with varying computational budgets. From each model, given a validation set of $n$ prompts $\mathcal{X}$, we obtain 4 prediction sets $\mathcal{Y} = \sigma(T(\mathcal{X}))$, $\mathcal{Y}_{12} = \sigma(T_\text{ULS, 12}(\mathcal{X}))$, $\mathcal{Y}_{8} = \sigma(T_{\text{ULS, 8}}(\mathcal{X}))$, $\mathcal{Y}_{4} = \sigma(T_{\text{ULS, 4}}(\mathcal{X}))$ from the Alpaca dataset. 
Each prediction set item consists of the full sequence $\bv{y}_{i}$ generated by the compressed model for a prompt $\bv{x}_i \in \mathcal{X}$.  The oracle assigns a cost $c \in \{4,8,12,24\}$ for each sequence to optimize the ROUGE-L score given a budget $\beta$ on average cost. See Appendix \ref{appendix:oracle_layer_skipping} for details and discussion of the oracle's sampling advantage.

\paragraph{Results:} Figure \ref{fig:oracle_stacked} illustrates the potential advantage of sequence-level computation allocation. Despite $T_{\text{ULS, 4}}$ achieving an average ROUGE-L score of $0.249$ compared to $0.342$ for the full model, the oracle can match the full model's performance using as few as 5.6 layers, selecting $T_{\text{ULS, 4}}$ $69.6\%$ of the time. The steep performance increase of the oracle illustrates how highly compressed models can be suitable for a majority of sequences, with the larger models rarely required. Even when given a budget that permits usage of all 24 layers for every sequence, the oracle still finds that the compressed models are preferred for more than $68\%$ of sequences. 

\vspace{-0.2cm}
\section{Conclusion and Limitations}
\vspace{-0.2cm}
 Our experimental results offer new guidance for the development of dynamic LLMs. We provide new evidence that layer skipping is a better alternative than early exiting for autoregressive transformer tasks. We highlighted the need for further research in skip controller design, as we found that the hidden states information is too complex for a simple controller. 
Finally, we provided a motivating experiment for a new approach: controlling layer skipping at the sequence level. The key limitation of our work is the limited scope of architectures and datasets, imposed by computational constraints. We made an effort to select models and tasks that are generally representative, but our findings may not generalize to other architectures and tasks. Additional limitations are discussed in Appendix \ref{appendix:limitations}. 

\section*{Acknowledgments}
We acknowledge the support of the Natural Sciences and Engineering Research Council of Canada (NSERC).

Cette recherche a été financée par le Conseil de recherches en sciences naturelles et en génie du Canada (CRSNG).

\bibliographystyle{IEEEtranN}
\bibliography{ref}

\newpage
\appendix

\section{Appendix}

\subsection{Related Work} \label{sec:related-works}
\subsubsection{Sample-agnostic approaches}
We deem approaches {\em sample-agnostic} if they use a fixed schedule to bypass certain model layers, independent of the samples being processed. In this category, \citet{layerdrop_fan} and \citet{unified_layer_skipping_liu} train encoder-decoder and decoder-only models, respectively, to use a constant skipping schedule for all sequences. Both observe that uniformly spacing the skipped layers achieves superior downstream performance over skipping consecutive layers. 

\subsubsection{Sample-aware approaches}
Sample-aware approaches use information about the samples provided to determine the computational budget to allocate. \citet{depth_adaptive_transformer_elbayad} propose an early exit strategy for encoder-decoder models, which consists of executing only the first $n$ layers and bypassing the rest of the model using an early exit classifier, based on the sample being computed. They show that learning a binary controller at each layer using the current hidden state achieves the best performance. \citet{calm_schuster} report similar findings using the hidden states on a summarization task, performing on par with a softmax thresholding approach, described in Appendix \ref{appendix:softmax_thresholding}. For decoder-only models, \citet{consistent_ee_zeng} learn a policy network for early exiting based on the intermediate hidden states. In contrast, \citet{zeng2023_learning_to_skip} learn a layer skipping controller which uses the hidden states to choose to skip a given layer. While the hidden state-based methods do outperform the fully random or static baselines presented, they do not make comparison to a learnable, token-agnostic method similar to the baseline in Section \ref{sec:hidden_state_evaluation}. This leaves some ambiguity regarding how much of the performance improvement is due to leveraging the context of a specific token compared to just learning the average importance of each layer.

\subsubsection{Routing with intermediate classifiers} \label{appendix:softmax_thresholding}
In early exit, a common method to determine the exit layer is to perform inference on intermediate classifiers after every layer, and to define a threshold on the classifier output distribution to trigger the early exit \citep{xin-etal-2020-deebert, liu-etal-2020-fastbert}. Although successful in other tasks, such an approach is especially costly in an NLG setting where the the large output vocabulary size requires a large linear classifier. Evaluating this classifier every layer adds unwanted overhead costs, which motivated research towards lightweight binary classifiers based on the hidden state \citep{depth_adaptive_transformer_elbayad, calm_schuster}. 

\subsubsection{KV cache management}
During autoregressive inference, the self-attention mechanism in each layer requires the key-value pairs of previous tokens, which are usually stored in a KV cache to avoid redundant computation. However, the key-value pairs are never computed when a layer is bypassed through early exiting or layer skipping. \citet{depth_adaptive_transformer_elbayad} identify this concern and propose to copy the hidden state of the closest previously executed layer as an approximation for the missing hidden state, and fill the missing KV cache entries using the key and value projections matrices of each layer. The same copying principle is used by \citet{zeng2023_learning_to_skip} for sample-aware layer skipping. \citet{bae-etal-2023-fast} propose to instead compute the exact KV cache entries when they become required, taking advantage of parallel computation with the current token. With this method, early exiting does not reduce the total computational cost of the model, but does reduce the inference latency by deferring and parallelizing the execution of skipped layers.

\subsubsection{Successes of sample-aware routing}
Intermediate hidden states in LLMs are high dimensional, complex embeddings which encode contextual information for Transformer blocks to process, so re-purposing them for tasks such as early exit prediction is a challenging undertaking in efficient dynamic inference. Some success has been observed in encoder-decoder models by \citeauthor{depth_adaptive_transformer_elbayad} and \citeauthor{calm_schuster}, but this architecture possesses a distinct advantage in the context of early exiting: Since the initial input is first encoded by the encoder, the early decoder layers have access to a complete encoding of the input, which is not available in a decoder-only architecture \citep{skipdecode_delcorro}. In the related field of Mixture of Experts (MoE), a linear gating network is used to route samples to a subset of experts using the hidden state with considerable success \citep{Shazeer-MoE, jiang2024mixtralexperts}. In this setting, the goal of the gating network is to balance the use of each experts while allowing them to specialize on a subset of the hidden state space. We conjecture that a linear gating network is sufficient to process the hidden states because it needs only to consistently and fairly divide inputs among experts, which are jointly trained to specialize on the subset of inputs assigned to them. In contrast, a binary controller in efficient dynamic inference must learn to identify inputs which do not require full computation for the next token to be successfully predicted. This involves predicting the output of decoder layers with several orders of magnitude more parameters than itself, using a hidden state which has not been trained to explicitly encode such information. 

\subsection{Datasets and model details} 
\label{appendix:datasets_and_models}
\subsubsection{Datasets}
\paragraph{Alpaca \citep{alpaca}}
All results in the main body of the paper are reported for the Alpaca dataset for the following reasons. First, it is a common NLG benchmark for instruction-following and long-form question answering \citep{chang2024survey}. Furthermore, the dataset has short prompts and long generation requirements, implying that the autoregressive generation phase dominates over the prompt decoding, which is the main target use case for layer skipping and early exiting. The dataset consists of 52,000 instructions and demonstrations used for instruction-tuning. The median prompt length is 43 tokens, while the median label length is 42 tokens. The maximum context length used throughout training is 512, and the split used throughout all experiments is 70\% train, 15\% validation, and 15\% test.

\paragraph{CNN-DM \citep{cnndm}}
Plots resulting from the use of the CNN-Daily Mail dataset are included in the appendix to further validate our findings on another popular NLG benchmark. The dataset consists of just over 300,000 unique news articles and associated summaries. The median prompt length is 788 tokens, and the median label length is 66 tokens. The maximum context length used for training is 2048, and the split used for this dataset is approximately 92\% train, 4\% validation, and 4\% test.

\subsubsection{Fine-tuned model training}
We use the OPT-decoder-only architecture \citep{zhang2022opt}, and perform standard fine-tuning on Alpaca and CNN-DM starting from the instruction-tuned checkpoint of OPT-1.3B by \citet{iyer_optiml}. All experiments are performed using 8X Nvidia P100 as compute. For Alpaca, we fine-tune for 6 epochs with early stopping and a patience of 1. The best evaluation ROUGE-L score is obtained after 3 epochs. We use a learning rate of $5e^{-5}$ which decays linearly over the training procedure. We use a batch size of 64, and an AdamW optimizer.

For CNN-DM, we fine-tune for 2 epochs with early stopping and a patience of 1. The best evaluation is obtained at 2 epochs. The same learning rate and optimizer is used with a batch size of 48. In both cases. the prompt is masked away from the loss, so the model is only trained on the completion task.

\subsection{Hidden state similarity details}
\label{appendix:hidden_state_similarity}
\subsubsection{Uniform layer skipping formal definition} \label{app:ULS_strategy}
We define the uniform layer skipping network with cost $c$ out of a total of $L$ layers as $T_{\text{ULS, c}}$. The execution of layer $l$ is determined by $G^l := U(l,c)$ using the recursion 
\begin{equation} \label{eq:ULS}
    U(l,c) = \ind \left( \left( \sum_{k<l} U(k,c)\right) \leq (l-1)\frac{c}{L} \right), U(1,c) := 1.
\end{equation}

\begin{figure}[!h]
    \centering
    \includegraphics[width=0.5\linewidth]{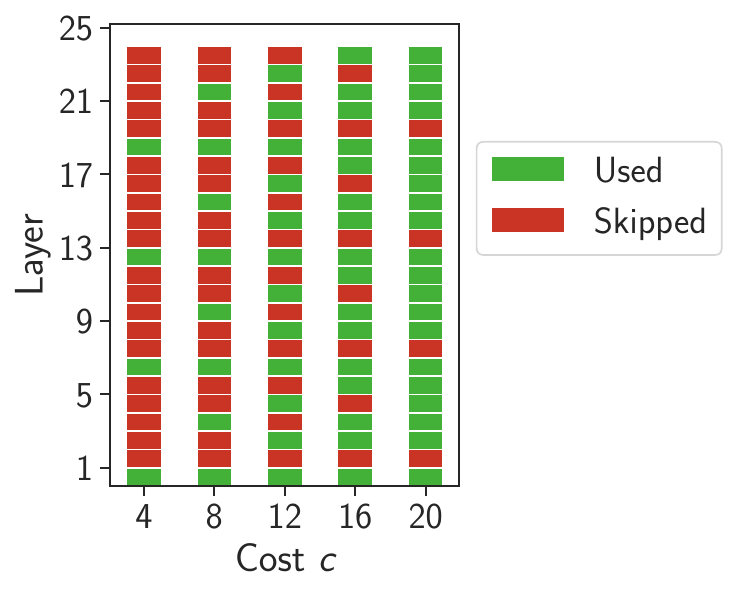}
    \caption{Uniform layer skipping strategy of a 24 layer model following Equation \ref{eq:ULS} for different computational cost $c$.}
    \label{app:fig:ULS_skip_distribution}
\end{figure}

Figure \ref{app:fig:ULS_skip_distribution} illustrates the produced layer skipping pattern for different computational costs $c$. 

\subsubsection{Random layer skipping when executing the first layer is not enforced}\label{app:RLS_without_L1}
\begin{figure}[!h]
    \centering
    \includegraphics[width=0.5\linewidth]{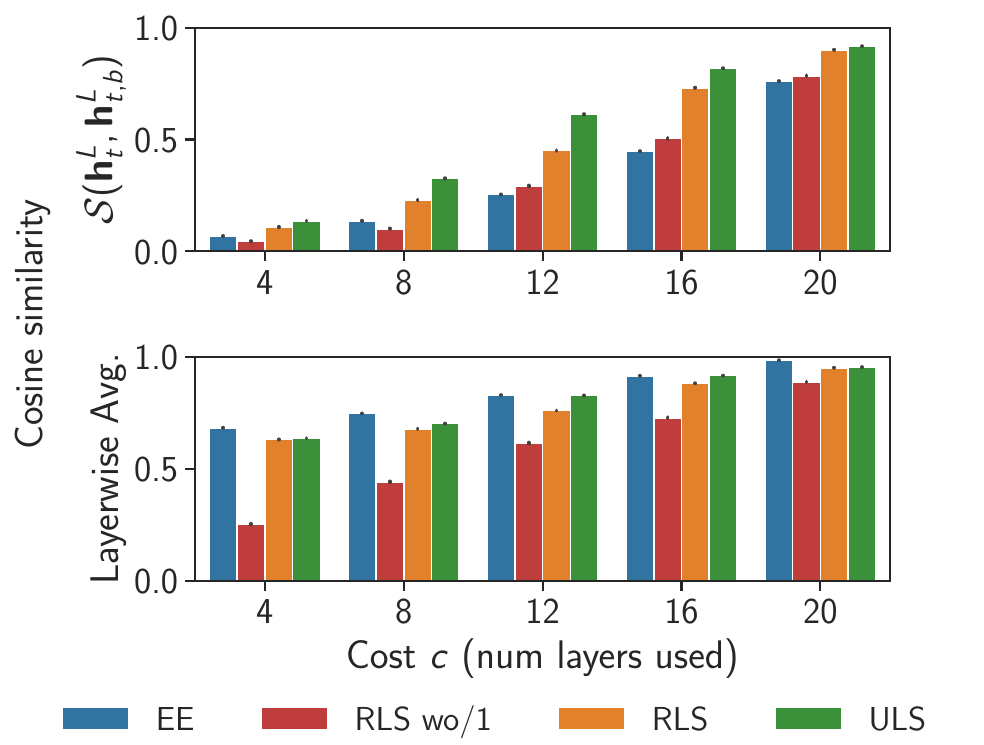}
    \caption{Using OPT-1.3B (24 layers) on the Alpaca dataset. We report the mean and 95\% confidence interval on the test set. Average cosine similarity of the final hidden states $\bv{h}^L_{t}$ (top) and layerwise intermediate hidden states $\bv{h}^l_{t}, 1\leq l \leq 23$ (bottom) obtained from each dynamic route strategy compared to the full model execution. We include RLS wo/1, random layer skipping without enforcing the execution of layer 1, and see that it performs consistently worse that RLS.}
    \label{app:fig:RLS_layerwise_and_final_similarities_alpaca}
\end{figure}

To justify the rule to enforce the execution of layer $l=1$ for all methods in Section \ref{sec:Comparing-EE-LS}, we include in Figure \ref{app:fig:RLS_layerwise_and_final_similarities_alpaca} the results of $T_{RLS wo/1}$, random layer skipping when layer 1 is permitted to be skipped. We observe that the performance is consistently worse than $T_{RLS}$, confirming the importance of executing the first layer in a frozen model.

\subsubsection{CNN-DM results for hidden state similarity}
\begin{figure}[!h]
    \centering
    \includegraphics[width=0.5\linewidth]{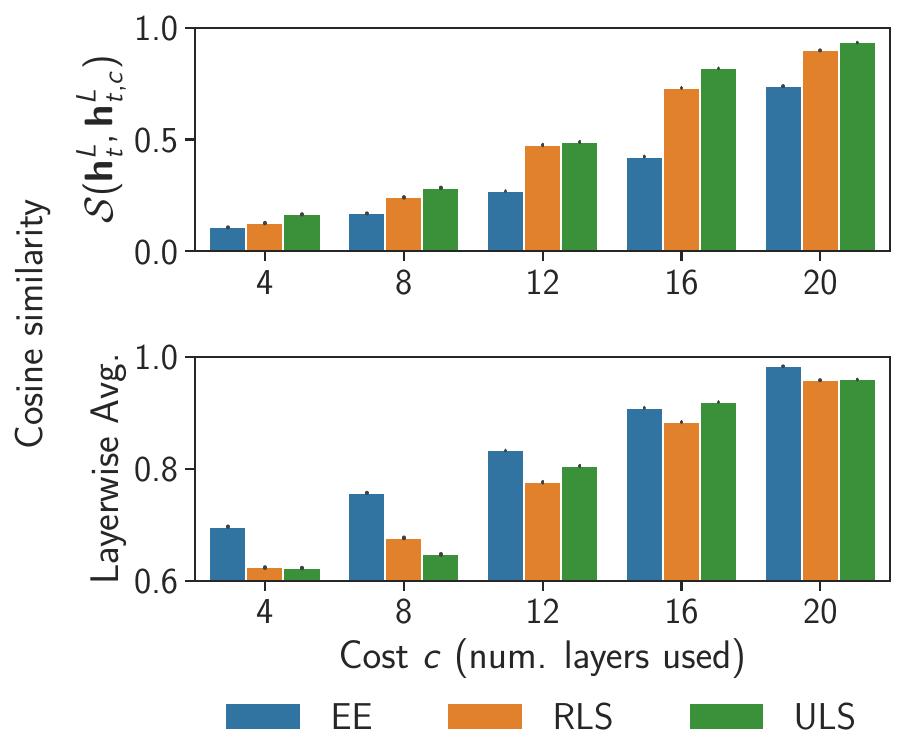}
    \caption{Using OPT-1.3B (24 layers) on the CNN-DM dataset. We report the mean and 95\% confidence interval on the test set. The cosine similarity of the final hidden states $\bv{h}^L_{t}$ (top) and layerwise intermediate hidden states $\bv{h}^l_{t}, 1\leq l \leq 23$ (bottom) obtained from early exit (EE), random and uniform layer skipping (RLS/ULS) is shown compared to the full model execution.}
    \label{app:fig:cnndm_cosine_similarity_vs_computational_budget}
\end{figure}

We present in Figure \ref{app:fig:cnndm_cosine_similarity_vs_computational_budget} a repetition of the Section \ref{sec:Comparing-EE-LS} experiment on the CNN-DM dataset. We observe a similar pattern, with the exception of early exiting having a higher cosine similarity of intermediate hidden states than uniform layer skipping at cost $c=12$.
\subsection{Hidden state versus fixed input layer skipping details} 

\subsubsection{Training setup} \label{appendix:training_setup_hidden_states}
We start with a standard fine-tuned model following Appendix \ref{appendix:datasets_and_models}. 
 After the fine-tuning is complete, the skip controller training phase begins and follows the procedure of \citet{zeng2023_learning_to_skip}. Controllers -- consisting of a single linear layer followed by a Gumbel softmax \citep{gumbel_softmax_repar} --  are added to every layer of the fine-tuned model except the first and last, and are jointly fine-tuned with the backbone. The KV-cache is propagated following the approach described in the CALM paper \citep{calm_schuster} and the prompt is propagated through the full network. We train the controller-augmented network for a maximum of 3 epochs, running evaluation every $\frac{1}{3}$ epoch and using early stopping with a patience of 1. The fixed input Gumbel training uses the same parameters as the hidden state Gumbel training.

\subsubsection{Per-layer skip ratios of the hidden state and fixed input skipping approaches}
\label{appendix:extra_hidden_versus_fixed}

\begin{figure}[!h]
    \centering
    \includegraphics[width=0.7\linewidth]{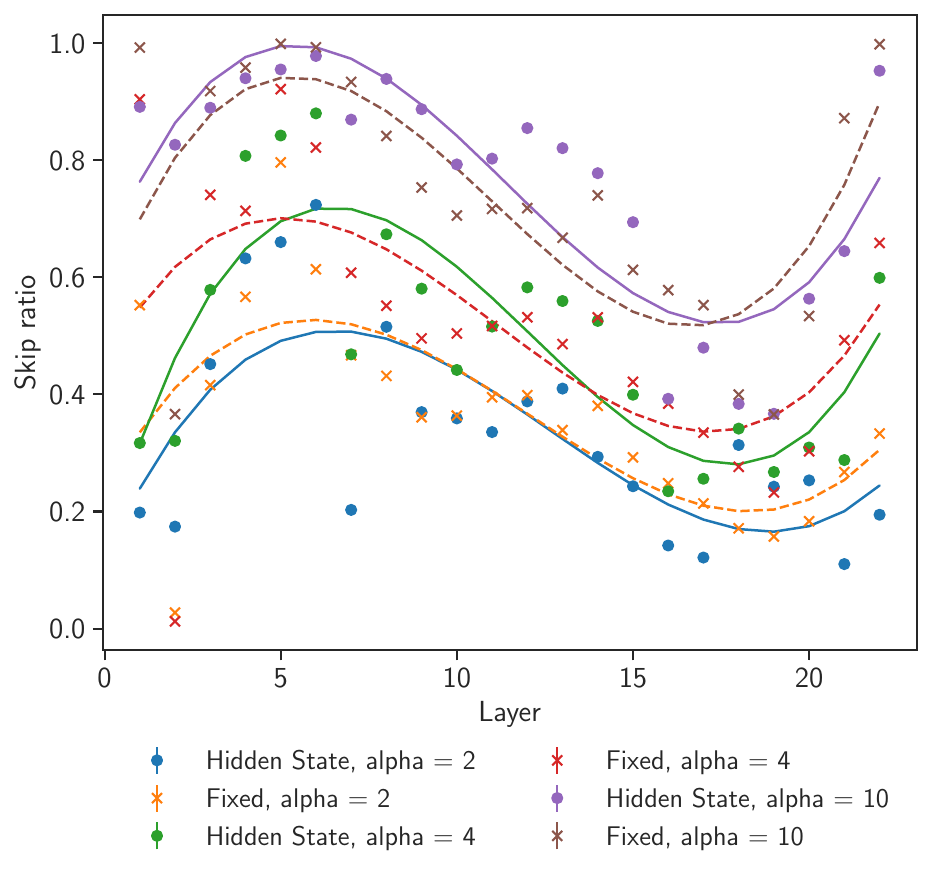}
    \caption{Per-layer skip ratios as a function of $\alpha$ and the layer number when using the hidden state or a fixed tensor as input to the skip controllers. The two approaches have very similar trends across different values of $\alpha$, showing that the learnt skip strategy depends more on the layer than the token.}
    \label{fig:per_layer_skip_ratios}
\end{figure}

In this section, we provide additional information about the experiment of Section \ref{sec:hidden_state_evaluation}. This experiment showed that there was no significant performance difference when using the hidden state compared to a fixed input to the skip controllers. In order to further investigate the skipping strategies adopted by each of the approaches, we plot the percentage of skips across all generated tokens for varying values of $\alpha$. The results are displayed in Figure \ref{fig:per_layer_skip_ratios}.

The results show very similar trends for both the hidden state and fixed input approaches across different values of $\alpha$, showing again that the two models learn a similar skipping strategy on average per layer.

\subsubsection{Ablation study of controller performance on token hidden states} \label{appendix:ablation_hidden_state}
\begin{figure}[!h]
    \centering
    \includegraphics[width=0.5\linewidth]{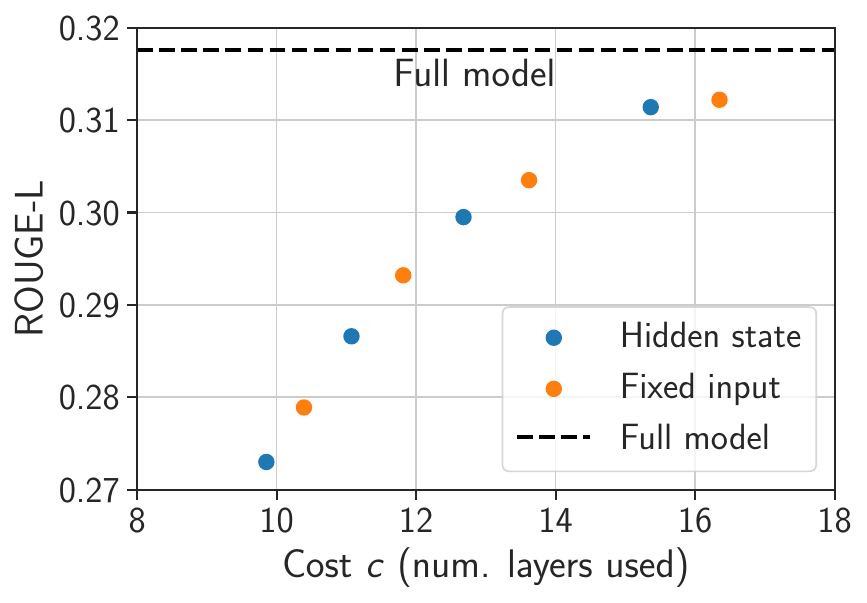}
    \caption{Comparison of cost-performance curves on  OPT-350M \citep{zhang2022opt} when training skip controllers to use the hidden states $\bv{h}^l$ or a fixed input. For this ablation, we fix the fine-tuned backbone to prevent distribution shifts in $\bv{h}^l$. We see that using the hidden state brings no significant advantage over the fixed input.}
    \label{app:fig:hs_gumbels_ablation}
\end{figure}

The surprising results of Section \ref{sec:hidden_state_evaluation} prompted us to further investigate why skip controllers using a token's hidden state $\bv{h}^l$ perform on-par with controllers fed a fixed input when using the loss formulation in Equation \eqref{eq:ce_loss}. One plausible explanation is that training the backbone jointly with the controllers as in \citet{zeng2023_learning_to_skip} causes a distribution shift in $\bv{h}^l$ making the controller learning task more challenging. \\
To test this hypothesis we first fine-tune the backbone with random layer dropout following \citet{layerdrop_fan}, freeze the backbone and subsequently train the controllers following Equation \eqref{eq:ce_loss}. As in Section \ref{sec:hidden_state_evaluation}, we compare the cost-performance trade-off between hidden-state and fixed input training for varying values of $\alpha$. We perform this experiment on OPT-350M \citep{zhang2022opt} trained on the Alpaca dataset. OPT-350M consists of 24 layers. As before, we always execute the first and last layers. We use a batch size of 20 and a maximum sequence length of 512. Figure \ref{app:fig:hs_gumbels_ablation} shows that even in this setting where the backbone is frozen, the controllers fail to extract meaningful information from $\bv{h}^l$ with the exception of the high cost regime (around 16 layers).

\subsection{Oracle algorithm details} \label{appendix:oracle_layer_skipping}

In this section, we provide additional information about the experiment of Section \ref{sec:oracle-layer-skipping}. Given the 4 prediction sets $\mathcal{Y} = \sigma(T(\mathcal{X}))$, $\mathcal{Y}_{12} = \sigma(T_{12}(\mathcal{X}))$, $\mathcal{Y}_{8} = \sigma(T_{8}(\mathcal{X}))$, $\mathcal{Y}_{4} = \sigma(T_4(\mathcal{X}))$ obtained by using the 24 (full), 12, 8 and 4-layer cost uniform layer skipping OPT-1.3B respectively, the oracle algorithm assigns to each input $\bv{x}_i$ a model $T_c$ so as to maximize the average performance metric (ROUGE-L score) given a maximum average computational budget $\beta$.

The problem can then be framed as a multiple choice knapsack problem \citep{knapsack} where each sequence $\bv{x}_i$ has to be assigned to exactly one of four models $T_c, c \in \{4,8,12,24\}$ to maximize average ROUGE-L score subject to a cost constraint. The solution to the algorithm for budget $\beta$ is an assignment of costs $\bv{S}^\beta_{1:n} \in \{4,8,12,24\}^n$, where $S^\beta_{i}$ denotes the cost $c$ of the model $T_c$ selected for input $\bx_i$. We can write the oracle objective $P$ as maximizing the average ROUGE-L score obtained given $\beta$:
\begin{equation} \label{eq:knap_oracle_budget}
   \begin{gathered}
    P = \argmax_{\bv{S}^\beta_{1:n}} \,\, \frac{1}{n}\sum_{i = 1}^n ROUGE(T_{S^\beta_{i}}(\bv{x}_i))) \\
    \text{subject to}\quad  \frac{1}{n}\sum_{i = 1}^n S^\beta_{i} \leq \beta\\
\end{gathered} 
\end{equation}
where $n=7800$ is the total number of samples considered. Using dynamic programming, we obtain the optimal assignment $\bv{S^*}^\beta_{1:n}$ and report the average ROUGE-L score $P$ for different $\beta$ in Section \ref{sec:oracle-layer-skipping}.

\subsubsection{Quantifying the sampling advantage} \label{appendix:oracle-sample-boost}
Although $\bv{S^*}^\beta_{1:n}$ is a valid solution achievable in theory by a perfect controller module, its advantage over the ULS models $T_4,T_8,T_{12}$ and $T$ cannot be solely attributed to the oracle assigning different computational paths based on a sequence's difficulty, as was hypothesized in Section \ref{sec:oracle-layer-skipping}. 
Because of the perfect information available to the oracle, there is an intrinsic sampling advantage present due to the natural variance in the ROUGE-L scores of each model. Even though the output quality of each model is correlated to its cost on average, stochasticity from the model's training and inference mean that, for all input $\bx_i$, we cannot assume that  
\begin{align*}
    ROUGE(T_{4}(\bx_i))\leq ROUGE(T_{8}(\bx_i))\leq ROUGE(T_{12}(\bx_i))\leq ROUGE(T(\bx_i)). 
\end{align*}

Generally, the advantage obtained in expectation from taking the maximum of multiple samples $\mathbb{E}[\max (X_1,X_2,\dots,X_k)] - \mathbb{E}[\frac{1}{k}\sum^k_{i=1}X_i]$ depends on the nature of the distributions, their variance and the number of samples considered. We refer to this advantage as the sampling advantage. As part of maximizing the objective $P$ in Equation \ref{eq:knap_oracle_budget}, the oracle will always choose the maximally performing model whose cost is within the budget available for a given input. The variance in the ROUGE-L scores of each model therefore introduces a sampling advantage, which in our case depends on the arbitrary number of models considered for the experiment. 

Although it is not feasible to construct an oracle benefiting only from the dynamic budget allocation advantage, we instead propose a greedy oracle which only benefits from sampling advantage to serve as a comparative baseline to the original oracle, in an effort to better approximate the dynamic budget allocation advantage. As suggested by its name, the greedy oracle is given perfect information regarding the performance of each model on each sequence, but finds an assignment $\bv{S}^\beta_{1:n}$ by selecting the locally optimal ${S}^\beta_{i}$ for each input, according to the objective
\begin{equation} \label{eq:knap_oracle_greedy}
   \begin{gathered}
    P = \frac{1}{n}\sum_{i = 1}^n \argmax_{{S}^\beta_{i}} ROUGE(T_{{S}^\beta_{i}}(\bv{x}_i))) \\
    \text{subject to}\quad S^\beta_{i} \leq \beta \quad \forall i.\\
\end{gathered} 
\end{equation}
Given an average computational budget $\beta$, the greedy oracle equally distributes the budget to all sequences. Unlike the original oracle which can reduce the budget given to one sequence to increase it for another who needs it more, the greedy oracle makes an isolated decision for every input, and therefore has no dynamic budget allocation advantage. However, by selecting the best model within the budget for each input, it is given a sampling advantage. While this sampling advantage may not perfectly align with that of our initial oracle, it serves as a reasonably close approximation. 

\begin{figure}[!h]
    \centering
    \includegraphics[width=0.55\linewidth]{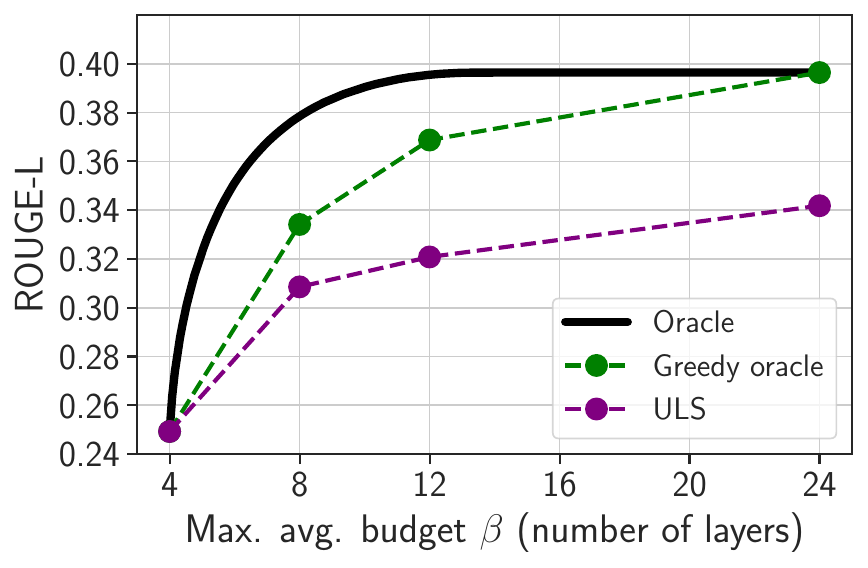}
    \caption{ROUGE-L of ULS models $T_{\text{4}}$, $T_{\text{8}}$, $T_\text{12}$ and $T$  compared to an oracle skip controller which selects the optimal model per sequence given an maximum average budget $\beta$ in number of layers. The greedy oracle cannot trade off increased budget for some sequence at the expense of others, but selects the model within the budget with the best performance.}
    \label{app:fig:weak_oracle}
\end{figure}

We evaluate the greedy oracle at $\beta = 4,8,12,24$, which corresponds to selecting the best model from $\{T_4\},\{T_4,T_8\},\{T_4,T_8,T_{12}\}$ and $\{T_4,T_8,T_{12},T\}$ respectively for each input. The performance difference shown in Figure \ref{app:fig:weak_oracle} between the ULS and greedy oracle plots corresponds to the sampling advantage obtained by selecting the best of the models with layer cost $c \leq \beta$ for each sequence, instead of only choosing the most costly one. As expected, there is no advantage at $\beta = 4$, since only the $T_4$ model can be used. The advantage grows with $\beta$ as more models can be considered. We can then compare the original oracle performance to the greedy oracle for the approximate marginal advantage obtained by dynamic budget allocation. We see that at $\beta=24$, there is no marginal advantage since dynamically allocating the budget has no effect when there already is the full $c=24$ available to every sequence. Crucially, we observe that there remains a notable marginal advantage in the budget-constrained interval $4 < \beta < 12$. For example, at $\beta = 8$, the dynamic budget allocation advantage of the original oracle over the greedy oracle is greater than the sampling advantage of the greedy oracle over the individual ULS model. Furthermore, the dynamic budget allocation advantage alone is sufficient for the oracle to outperform the full model using only $\beta = 8$, which validates the hypothesis that there is potential to achieve impressive computational savings from sequence-level budget allocation.

\subsubsection{Accounting for the generation length in the computational cost} \label{appendix:oracle-seq-length}
\begin{figure}[!h]
    \centering
    \includegraphics[width=0.55\linewidth]{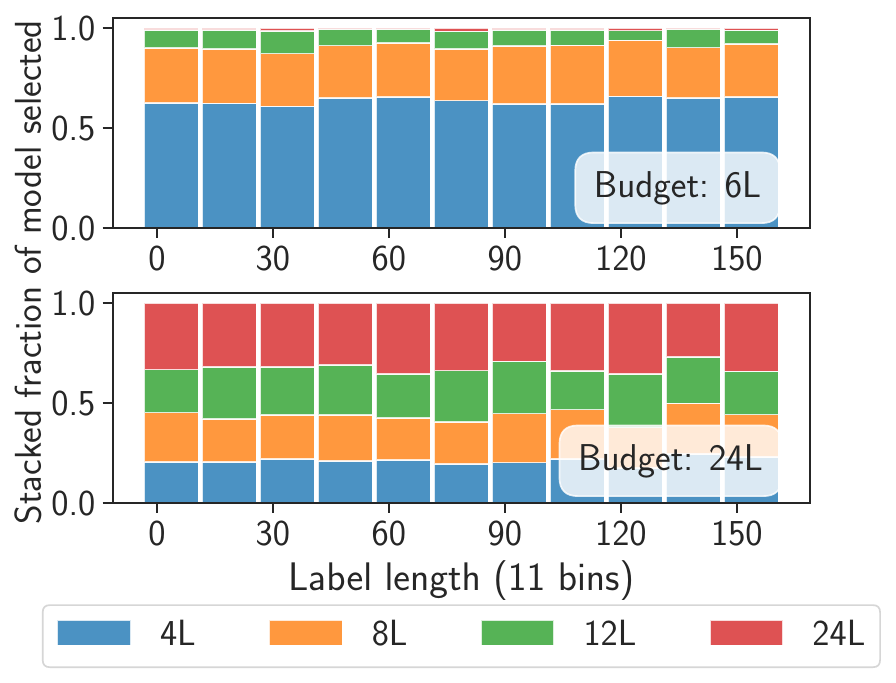}
    \caption{OPT-1.3B, Alpaca: Network selection for binned label lengths for average budgets $\beta$ of 6 (top) and 24 (bottom). We see that there is no significant variation of network selection distribution across bins.}
    \label{app:fig:label_length_bins_2}
\end{figure}

We note that the second line of Equation \eqref{eq:knap_oracle_budget}, which measures the total computational cost over the entire dataset, does not account for the length the label for $\bv{x}_i$ and only depends on the number of layers of the network selected for $\bv{x}_i$. Ignoring the label length allows to significantly speed up the knapsack solver by reducing the search space. Nonetheless, we verify that the model selection is independent of the label length by performing a chi-squared test of homogeneity at two different average budgets. Specifically, we bin the label lengths into 11 bins $\mathcal{B}_{i \in \{1, \dots, 11\}}$ of width 15. That is, $\mathcal{B}_i$ corresponds to generations whose length is in $[15(i - 1) + 1, 15i]$ for $i \in \{1, \dots, 10\}$. The $11^{th}$ bin contains all sequences whose label length is above 150. For a given average budget $\beta$, we obtain the proportion of samples routed on each compressed network for each bin $\mathcal{B}_i$ denoted by $p_{\beta, 24, \mathcal{B}_i}$, $p_{\beta, 12, \mathcal{B}_i}$, $p_{\beta, 8, \mathcal{B}_i}$, $p_{\beta, 4, \mathcal{B}_i}$. We then conduct a chi-square homogeneity test to determine whether these proportions significantly vary from one bin to another. For $\beta = 6$, corresponding to the yellow star in Figure \ref{fig:oracle_stacked}, we obtain $\chi^2=29.02$ and a p-value of 0.51. We conclude that there isn't sufficient evidence to reject the null-hypothesis that the network selection across bins follows the same distribution using a significance level $\alpha=0.05$. For $\beta = 24$, we obtain $\chi^2=40.57$ a p-value of 0.09. Similarly, we conclude that there isn't enough evidence to reject the null hypothesis at $\alpha=0.05$. Figure \ref{app:fig:label_length_bins_2} provides a visualization of the distribution of network selections across bins for these two average budgets.

\subsubsection{CNN-DM results for the sequence-level oracle}
\begin{figure}[!h]
    \centering
    \includegraphics[width=0.55\linewidth]{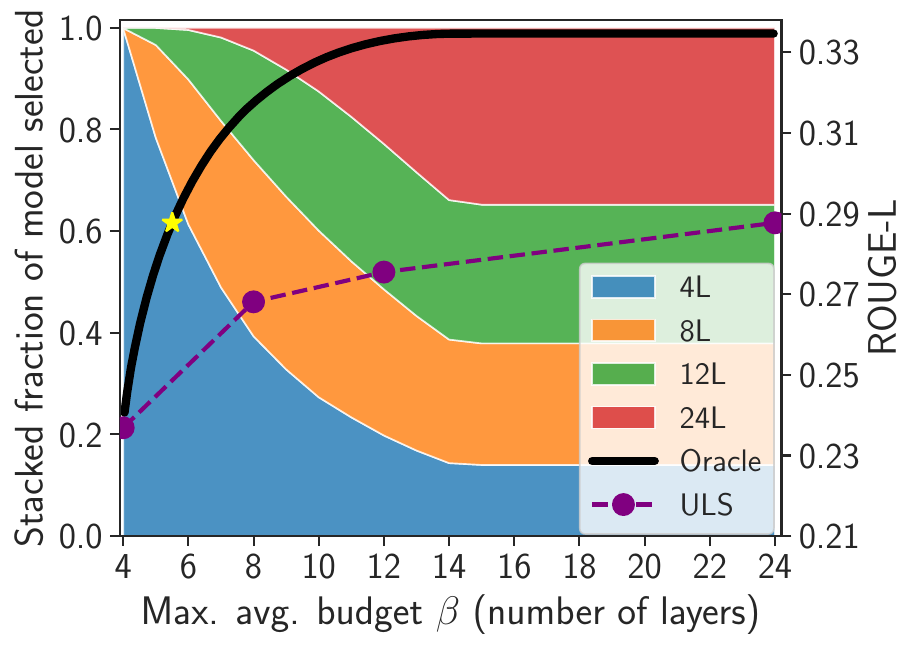}
    \caption{CNN-DM dataset: ROUGE-L (right axis) of  $T$, $T_\text{ULS, 12}$, $T_{\text{ULS, 8}}$, and $T_{\text{ULS, 4}}$ compared to an oracle skip controller that selects the optimal model per sequence given a global budget. For each global budget, the selection percentage of each model by the oracle is shown as a stacked plot (left axis). The yellow star denotes where the oracle can match the performance of the full static model, using an average of 5.5/24 layers ($22.9\%$).}
    \label{app:fig:cnndm_oracle_stacked}
\end{figure}

We present in Figure \ref{app:fig:cnndm_oracle_stacked} the oracle results for OPT-1.3B trained on CNN-DM following the experimental setup in Section \ref{sec:oracle-layer-skipping}. The test set contains $n=11490$ prompts. We obtain a similar performance curve and model selection percentage to Alpaca, with the exception that the full model $T$ is used with a higher percentage relative to $T_4$ for large values of $\beta$. 

\subsection{Limitations} \label{appendix:limitations}

There are several limitations that should be considered when interpreting the results. First, due to computational constraints, the scope of architectures and datasets we used for our experiments is limited. We made an effort to select models and tasks that are generally representative, but our findings may not be broadly generalizable to different architectures and tasks. Second, the performance of our gate controller is likely influenced by its design, and we only presented one design architecture. Therefore, it is possible that a better gate controller design could potentially utilize the hidden state more effectively. However, we do stress that our design aligns with architectures used in previous work, and that the computational budget dictates that a simple controller must be employed.
Last, as the oracle used in our analysis  in Figure~\ref{fig:oracle_stacked} directly uses test ROUGE scores for assignment, it can select the best outcome from multiple samples, even if they originate from the same distribution. As a result, the oracle achieves far beyond the ``attainable performance'' and surpasses a Bayes optimal model. The goal of the experiment is to highlight that there is a sequence-level selection that achieves a dramatic performance benefit. This motivates the development of a technique that can make a decision based on the prompt, without the benefit of hindsight. While we cannot provide the exact Bayes optimal performance, in Appendix \ref{appendix:oracle-sample-boost} we approximated the effect of the sampling advantage and showed that there remains a notable improvement .


\end{document}